\documentclass[letterpaper, 10 pt, conference]{ieeeconf}  

\IEEEoverridecommandlockouts                              
\usepackage{graphics} 
\usepackage{amsmath} 
\usepackage{dsfont}
\usepackage{graphicx}
\DeclareMathOperator*{\argmax}{argmax} 

\title{\LARGE \bf
MergeNet: A Deep Net Architecture for Small Obstacle Discovery}

\author{Krishnam Gupta$^{1}$, Syed Ashar Javed$^{2}$, Vineet Gandhi$^{2}$ and K. Madhava Krishna$^{2}$
\thanks{$^{1}$Krishnam Gupta is with Microsoft, India. Work done as graduate student at International Institute of Information Technology Hyderabad (IIIT-H), India.
$^{2}$Syed Ashar Javed, Vineet Gandhi and K. Madhava Krishna are with Kohli Center of Intelligent Systems (KCIS), IIIT-H, India. 
   {\tt\small krishnam.gupta@microsoft.com},
   {\tt\small mkrishna@iiit.ac.in}}%
}

\begin{document}

\maketitle
\thispagestyle{empty}
\pagestyle{empty}

\begin{abstract}

We present here, a novel network architecture called MergeNet for discovering small obstacles for on-road scenes in the context of autonomous driving. The basis of the architecture rests on the central consideration of training with less amount of data since the physical setup and the annotation process for small obstacles is hard to scale. For making effective use of the limited data, we propose a multi-stage training procedure involving weight-sharing, separate learning of low and high level features from the RGBD input and a refining stage which learns to fuse the obtained complementary features. The model is trained and evaluated on the Lost and Found dataset and is able to achieve state-of-art results with just 135 images in comparison to the 1000 images used by the previous benchmark. Additionally, we also compare our results with recent methods trained on 6000 images and show that our method achieves comparable performance with only 1000 training samples. 


\end{abstract}

\section{INTRODUCTION}
\label{sec:intro}

The importance of small obstacle discovery for on-road autonomous driving cannot be overstated. Small obstacles such as bricks, stones and rocks pose a veritable hazard to driving, especially to the state estimation modules that are a core constituent of such systems. Some times these obstacles can take the shape of stray dogs and cats that are entailed protection. Many a time these objects are too low on the road and go unnoticed on depth and point cloud maps obtained from state of the art range sensors such as 3D LIDAR. The problem slowly seems to be generating interest in the robotic and vision community \cite{Carsten-IROS16, Carsten-NIPS17}, not without a reason. For one, even the best of range sensors such as 3D LIDAR can find segmenting obstacles of height 15-25cms from a distance of 10m or more rather challenging. The problem is more pronounced with low cost low baseline stereo rigs, wherein the disparity profile can hardly be used to discern such obstacles from the background when they are at a depth of 5m or more.

Introspection reveals that the problem is difficult to solve purely based on appearance cues even with the best of the state of the art deep convolutional networks since gradients in the image can be caused equally due to changes in appearance such as markings and zebra crossings on the road as much as it could be due to obstacle edges. This problem aggravates in case the obstacles are small. Hence, an apt combination of both appearance and depth or disparity evidences is more likely to perform the task better. Recent efforts~\cite{valada2017adapnet} on multi modal fusion also suggests likewise. 

Most of the previous works~\cite{Oniga-ITSC11, Carsten-IROS16, Suryansh-ICRA14} in small obstacle detection are based on low level image and depth profile analysis, which are prone to errors due to noise in depth computation (especially  while using a stereo rig). The challenge in naive application of recently successful deep learning architectures is the limited availability of annotated data. In this paper, we propose a novel deep learning architecture called MergeNet which can be trained using as low as 135 images to obtain state of the art results. 

\begin{figure*}[thpb]
     \centering
  \includegraphics[width=\textwidth,height=0.6\textwidth]{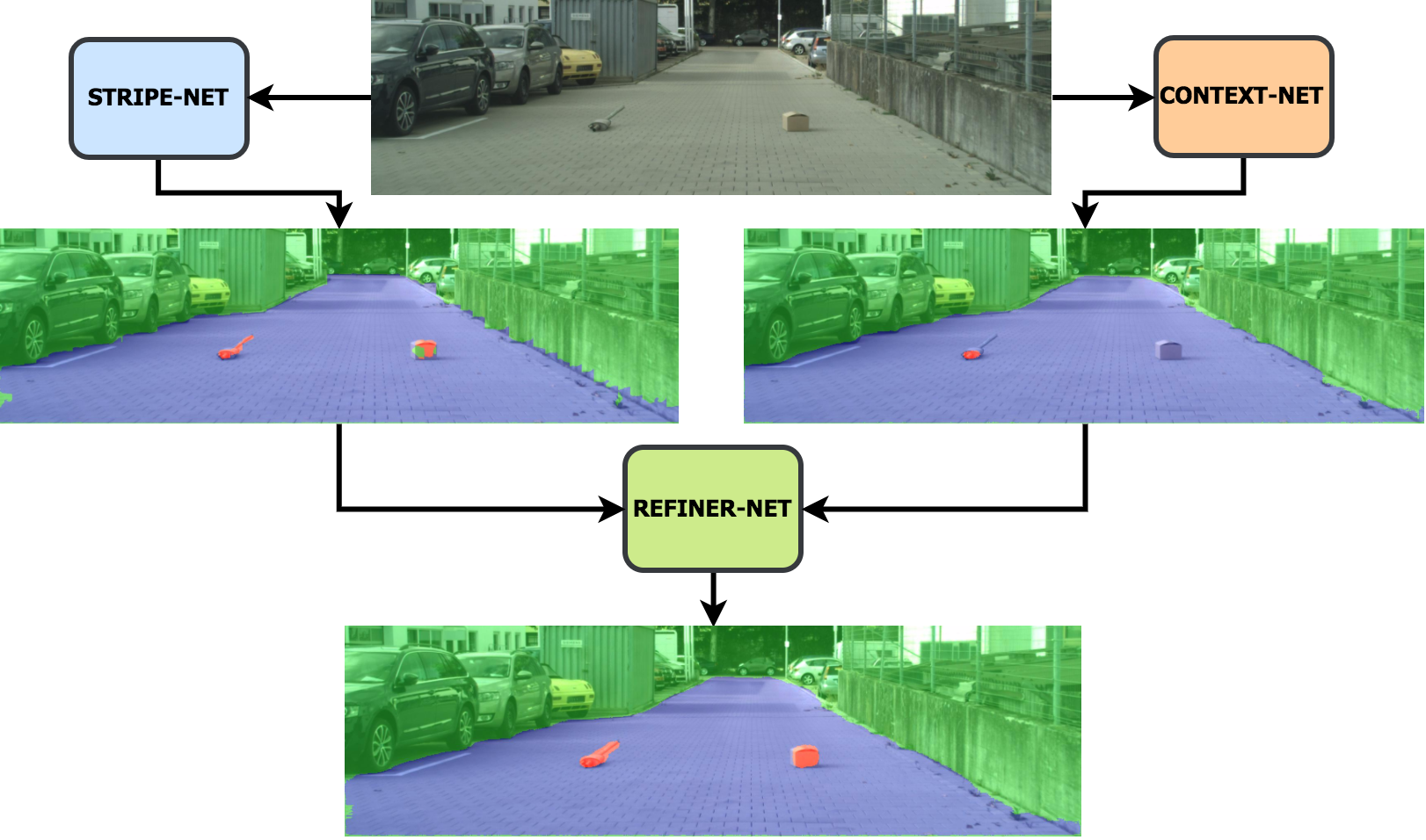}
     \caption{An overview of our network components and their outputs (the obstacle category is marked in red, the off road category is marked in green and the road category is marked in blue). The STRIPE-NET, which takes individual vertical stripes of RGBD data as input, is effective at detecting small objects (but with false positives and noisy road boundaries) whereas the CONTEXT-NET, which takes full RGB images as input, better preserves the overall structure of the scene (but misses out on small objects on the road). The REFINER-NET learns to combine the best of both worlds and outputs accurate and smooth segmentation maps.}
     \label{motivation}\vspace{-1em}
   \end{figure*}

We pose the problem of obstacle detection as that of segmenting the road scene into multiple classes. The proposed model consists of three key networks, namely the stripe-net, the context-net and the refiner-net. Stripe-net is a fully convolutional encoder-decoder model which is trained with column-wise strips of RGBD input (each training image is divided into a set of non overlapping vertical strips and fed to the network individually). The key idea behind Stripe-net is twofold: (a) learning discriminative features at a low-level by only attending to the vertical pathway of a road scene and (b) sharing parameters across the stripes to ensure lower model complexity and in turn reducing susceptibility to overfit even on small datasets. Context-net is also a fully convolutional encoder-decoder, but is trained on the full image. The role of this network is to incorporate global features which typically span a width higher than the stripe-width used in the previous network. Global coherence and important contextual cues are more suitably learnt using this network. Finally, the refiner-net is used for aggregating both the low and high level features and making the final prediction. Figure~\ref{motivation} illustrates a motivating example, showing the results at different stages of the proposed architecture. 

Formally, we make the following contributions:

\begin{itemize}

\item We propose a novel three-staged architecture for segmenting out the small on-road obstacles. The model design, the multi-modal data input and the fusion of features obtained at multiple spatial scales enable us to exploit the structure in a road scene while preserving the details necessary for small obstacle detection.

\item The proposed network can be efficiently trained for the task of semantic segmentation from as few as 135 images. Thus it makes a much needed effort in the direction of applying deep learning architectures in such data deficient applications. 

\item We test our model on the Lost and Found dataset \cite{Carsten-IROS16} and show an improvement of $19\%$ in the instance-level detection rate even when using a tenth of the training dataset and an improvement of around $30\%$ if we use the full dataset. We also achieve comparable results with \cite{Carsten-NIPS17}, while employing only one sixth of the training data.

\end{itemize}

The rest of the paper is organized as follows: Section~\ref{sec:related_work} lists the
related work. The proposed architecture is detailed in Section~\ref{sec:method}. The experiments and results are presented in Section~\ref{sec:experiments} and Section~\ref{sec:results}. The final section comprises of conclusions and future work.

\section{Related Work}
\label{sec:related_work}


Early efforts on small obstacle detection were limited to indoor scenes. Zhou and Baoxin~\cite{zhou2006robust} presented a solution for obstacle detection using homography based ground plane estimation algorithm. The work was extended in~\cite{kumar2014markov} for smaller obstacles by combining multiple cues like homography estimation, superpixel segmentation and a line segment detector into in a MRF framework. Early outdoor efforts on the other hand were focused on problems like curb detection~\cite{Homm-IV10, Oniga-ITS10}. One line of work~\cite{Homm-IV10} was based on probability occupancy maps~\cite{elfes1989using}, which is created by orthogonal projection of the 3D world onto a plane parallel to the road (assuming a structured environment where the floor surface is approximately planar). The plane is then discretized into cells to form a grid and the algorithm then predicts the occupancy likelihood of each cell. Another line of work~\cite{oniga2010processing, Oniga-ITS10} utilized digital elevation map (DEM's), which builds a height based cartesian occupancy grid and uses it for road surface estimation and obstacle detection.

Recent efforts~\cite{Carsten-IROS16,Carsten-NIPS17} have been made to extend the specific problem of small obstacle detection to outdoor settings for applications concerning autonomous driving and driver assistance. In~\cite{Carsten-IROS16} the Lost and Found dataset for small obstacle detection is presented along with three statistical hypothesis tests for detecting small obstacles. The extension of this work~\cite{Carsten-NIPS17} combines deep learning with hypothesis testing. A deep network termed UON which uses a Fully Convolutions Network~\cite{FCN} with GoogleNet architecture is used for obtaining a semantic segmentation map of the scene. The network is trained on 6000 RGB images, combining images from both Lost and Found dataset and the Cityscape dataset~\cite{cordts2016cityscapes}. The UON segmentation of the scene is then converted to a stixel realization~\cite{Frank-BMVC11} and fused with the hypothesis models of~\cite{Carsten-IROS16} to come up with the eventual segmentation. Our method on the contrary is void of any low/mid level post processing.

  \begin{figure*}[thpb]
     \centering
      \includegraphics[width=\textwidth]{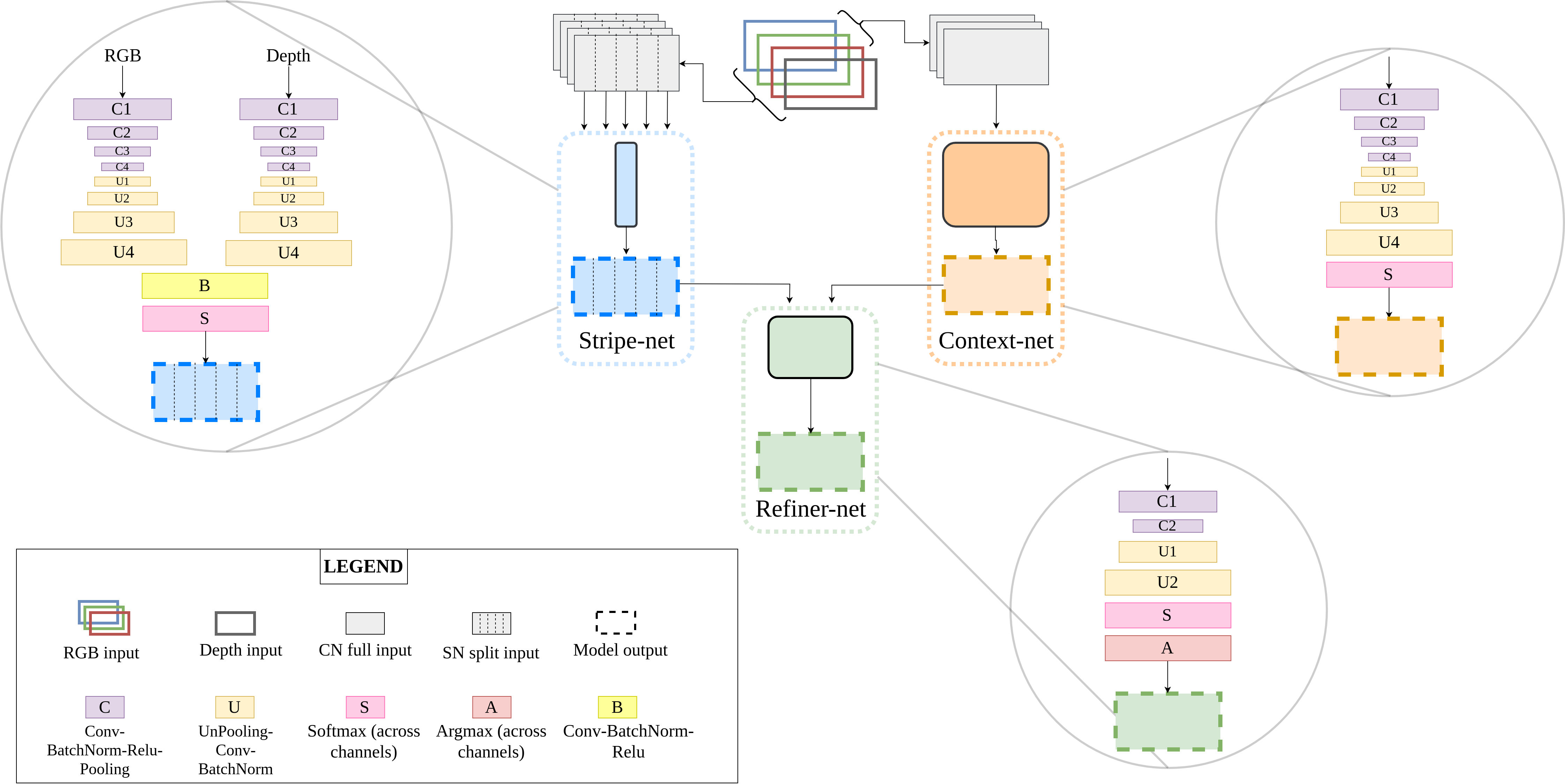}
     \caption{Model architecture is shown. The three components of the pipeline are magnified and illustrated in detail (best viewed in color)}
      \label{fig:model}\vspace{-1em}
   \end{figure*}
   
Since the small obstacle detection task can be expressed as that of semantic segmentation of the obstacles, it also makes sense to examine possible architectures pertaining to the segmentation literature. The typical semantic segmentation networks like ~\cite{farabet2013learning}, the Fully Convolutional Network ~\cite{FCN} and SegNet~\cite{badrinarayanan2015segnet} apply hierarchical and bottom-up convolutions and pooling to downsample an image, and then use layers like bilinear upsampling, unpooling, atrous convolution or deconvolution (strided convolution) to obtain a dense segmentation map. More advanced approaches like ~\cite{dai2015convolutional, sharma2015deep} utilize multi-scale or contextual features for learning this dense mapping. Although many competitive architectures for general application of semantic segmentation exist, replication of these models for specific tasks like obstacle detection yields inferior performance. Moreover, the labeled data required for these complex models also limits the direct applicability of standard semantic segmentation models, a problem which we try to resolve in our work.

\section{Learning to detect obstacles}
\label{sec:method}

Before formulating the problem and the model, it is useful to analyze the nature of the problem and the properties a good solution should possess. Firstly, the detection model should be capable of detecting small obstacles which are often far away on the road or don't appear like an obstacle (brick slabs of low height often resemble a cement road). Similarly, detection should preclude artifacts like zebra crossing and chalk markings on the road. Secondly, the effort required for the physical setup of constructing and annotating a small obstacle dataset for autonomous vehicles is considerable, so it makes sense to have a parsimonious learning model which isn't data intensive. Also, less training data makes a sufficiently parametrized model prone to overfitting, which is to be kept in check. Thirdly, the model should be deployable on an autonomous vehicle with respect to the memory and inference speed.

\subsection{Problem formulation}

The obstacle detection problem is posed as that of semantic segmentation of the video frames. The aim is then to learn a mapping $ f(w, x): X \rightarrow L$ for each of the $N$ pixels, where $x_i=(v_i, d_i) \in X$ are ordered pairs of visual RGB and depth input respectively and $l_i \in L$ are the set of labels for each pixel and $w$ are the parameters for the model being used. Three labels, namely \textit{'road'}, \textit{'off road'} and \textit{'small obstacle'} are considered. Due to the various challenges mentioned in the previous section, a staggered approach is used to learn the mapping $f$ as compositions of multiple functions. Each of these mappings is learnt with a loss function defined as the per-pixel cross-entropy loss, given as follows:

\begin{equation}
L(w) = -\sum_{i=1}^{N} \sum_{l=1}^{|L|} \mathds{1}{\{l=y_i\}}log(y_i)
\end{equation}

where, $\mathds{1}$ is the indicator function and $y_i \in Y$ is the softmax probability for pixel $i$. The final output is taken as

\begin{equation}
l_{predicted} = \argmax_{l \in L} (Y) 
\end{equation}

\subsection{The MergeNet model}
The core intuition behind the MergeNet model is that learning the low-level features of a road scene, combined with high-level features learnt from the context of the scene can jointly produce better segmentation maps. 
The MergeNet consists of three individual CNNs, each implemented as a fully convolutional network. The encoder and decoder blocks used for upsampling and downsampling of the image are similar to Segnet (basic version)~\cite{badrinarayanan2015segnet}, differing only in the number of layers and channels. Each encoder downsamples the input resolution through a series of convolution, batch normalization and pooling layers and then upsamples the encoded features though a series of deconvolution, batch normalization and unpooling layers, back to the original resolution. The first two networks learn at different levels of spatial abstraction while the third is a refiner network which exploits complementary information from the first two networks to produce a refined semantic map. These three models are denoted by functions $g_{stripe}(w_s, x_s)$, $g_{context}(w_c, x_c)$ and $g_{refine}(w_r, x_r)$ in the following sub-sections. Figure~\ref{fig:model} shows the detailed structure of the whole pipeline and its various modules.

\textbf{Stripe Network}
The stripe network is the model responsible for learning low-level features present within a narrow vertical band of the image. The whole image is split vertically into $k$ strips of equal width and the function $g_{stripe}(w_s, x_s)$ is learnt using a shuffled set of strips from the whole training dataset using ground truth labels of corresponding strips as supervision. 

The network consists of two parallel encoding and decoding branches, one for the RGB channel and another for the depth channel. Each branch contains four layers of convolutional downsampling in the encoder followed by 4 layers of upsampling in the decoder. These branches are subsequently fused across the channel dimension to obtain a combined map which is used for pixel-wise predictions of the three classes using a softmax layer.  

The use of vertical strips $x_s$ offers multiple benefits. It allows the model to concentrate on the discriminative features present only within the narrow vertical band. This, when combined with the context network, can be seen as a form of curriculum learning of features where the easier features are learnt first, followed by learning from more global and information-dense parts of the image. Secondly, the vertical strips allow the disparity map to contribute critical depth information. Within a narrow strip of a typical road scene, the disparity map follows an increasing trend of depth until an obstacle is encountered, the depth at which point, flattens out for some time before regaining its original trend. This can act as a very strong signal for the presence of obstacles. Thirdly, the use of strips allow the network to learn useful features with only a small set of parameters which are shared across all vertically split inputs. This is crucial for preventing overfitting on the small training dataset.

\textbf{Context Network}
The context network is responsible for learning complementary features with respect to the stripe network. The function $g_{context}(w_c, x_c)$ is learnt through full RGB images as input $x_c$, trained with the full image ground truth as supervision. The idea is to learn global context which can complement the stripe network. The context network contains the same number of layers and channels as that of any individual stream of the stripe network and is also similar in architecture to the fully convolutional networks generally used for semantic segmentation task. But the advantage of using it in conjunction with the stripe network is evident from the experimental results which highlight the limitation of the context network in detecting very small obstacles. Apart from providing the global features for obstacle detection, the context network also generates output segmentation maps which are smoother and more coherent with respect to neighboring pixels. Note that as opposed to stripe net, including the depth channel leads to reduced performance, which might be attributed to the use of complex RGBD data coupled with training with very few samples.

\textbf{Refiner Network}
The refiner network $g_{refine}(w_r, x_r)$ is learnt using $x_r$ = $(y_s, y_c)$ where $y_s$ and $y_c$ refer to the class normalized output features from $g_{stripe}(w_s, x_s)$ and $g_{context}(w_c, x_c)$ respectively. The input to this network is therefore the output maps of the previous networks concatenated across the channel axis. This model contains only two convolutional layers each for encoding and decoding as the task of segmentation of raw images is simplified to identifying relevant features from the previously trained outputs. The visualization of features from the stripe and context network underlines the need of a refiner network to learn from the complementarity of features.

The complete MergeNet model is a composition of the three networks:
\begin{equation}
f(w, x) = g_{r}(w_r, (g_{s}(w_s, x_s), g_{c}(w_c, x_c)))
\end{equation}

where $g_{s}$, $g_{c}$ and $g_{r}$ are the stripe, context and refiner models

\subsection{Implementation details}
The context model and refiner model are trained with images of resolution $256*896$ whereas the stripe network is trained with images of size $256*32$ by dividing each image into $k$ vertical strips where $k=28$. Thus each input is of size $256*32$. The strip-width of $32$ is chosen through cross-validation. Adam optimizer~\cite{kingma2014adam} is used while training with an initial learning rate of $0.01$. The batch size is taken as 4 for the context and refiner network and 32 for the stripe network. Training is continued until the model starts to overfit on the validation set. The context and refiner networks are trained with a weighted cross entropy loss to account for the difference in the number of pixels of each class in the dataset. The weights for each class are set as inversely proportional to the ratio of number of class pixels and the number of total pixels. The stripe network is not explicitly weighted at the loss level, but the shuffled stripes are sampled and fed to the network such that the expected value of occurrence of each class pixel is balanced out while training.

During inference, the individual networks are merged and a single segmentation map is produced by the refiner network. The individual strips of the input image $x$ can be passed through the stripe network in parallel as the weights are shared.

\section{Experiments}
\label{sec:experiments}
The following section details the evaluation of our model. 
\subsection{Dataset}
The Lost and Found dataset~\cite{Carsten-IROS16} is used for both training and testing of our network. The dataset consists of around 2200 frames of road scene obtained from 112 stereo videos, along with the pixel-level annotation of each frame pertaining to the road category, the off road category and the small obstacle category.  The dataset has a challenging test set of 1200 images which contains varying road illumination, different small objects present at long distances, non-uniform road texture, appearance and pathways and many non-obstacle class objects acting as distractors off the road. We train our model using two sets of training data, one with the complete training set of 1036 images and another with a reduced subset of only 135 images sampled equally across all the training image sequences. We evaluate our model on the released test set.

\subsection{Evaluation metrics}
We evaluate our model on both pixel-level and instance-level metrics. While choosing the metrics, there are two primary goals. The first is to ensure fair comparability of our results with the previous state-of-art methods, many of which employ some approach-dependent metric. The second is to assess the performance of our network through generic and approach independent metrics which can be used without any method-coupled adaptation in the future work.

\textbf{Pixel-wise detection rate}
Pixel-level detection rate (PDR) is defined as the fraction of pixels of the obstacle class, taken across the test set, which are correctly detected by the network. Formally, this metric is calculated as:

\begin{equation}
PDR = \dfrac{CDP_{obstacle}}{TP_{obstacle}}
\end{equation}

where $CDP_{obstacle}$ refers to the correctly detected pixels of the obstacle class and $TP_{obstacle}$ refers to the total pixels of the obstacle class.

\textbf{Instance-wise detection rate}
Instance-level detection rate (IDR) is defined as the fraction of obstacle instances, taken across the dataset, which are detected by the network. For this metric, an instance is marked correctly detected if more than $50\%$ of the pixels of the predicted obstacle overlaps with the ground truth of that instance. For extracting instances from pixel-level predictions, we convert the segmented map into a binary image with obstacles/ non-obstacle classes. Then, a 4-connectivity based connected component algorithm is used to obtain instance-level maps of just the obstacle class. The metric is formally calculated as:

\begin{equation}
IDR = \dfrac{CDI_{obstacle}}{TI_{obstacle}}
\end{equation}

where $CDI_{obstacle}$ refers to correctly detected instances of the obstacle class and $TI_{obstacle}$ refers to the total instances of the obstacle class, taken across the entire dataset.

\textbf{Pixel-wise false positives}
Pixel-level false positives (PFP) is defined as the fraction of pixels of the non-obstacle classes, taken across the dataset, which are incorrectly marked as an obstacle. Formally, this metric is calculated as:

\begin{equation}
PFP = \dfrac{IDP_{obstacle}}{TP_{non\_obstacle}}
\end{equation}

where $IDP_{obstacle}$ refers to the incorrectly detected pixels of the obstacle class and $TP_{non\_obstacle}$ refers to the total pixels of the non-obstacle class.

\textbf{Instance-wise false positives}
Lastly, instance-level false positives (IFP) is defined as the fraction of non-obstacle instances, taken across the dataset, which are incorrectly detected by the network. Similar to IDR, connected components algorithm is used to get instance-level predictions. The metric is formally calculated as:


\begin{equation}
IFP = \dfrac{IDI_{obstacle}}{d}
\end{equation}

where $d$ is the number of frames in the testing dataset and $IDI_{obstacle}$ refers to incorrectly detected instances of the obstacle class, taken across the entire dataset.

\section{Results}
\label{sec:results}
\begin{table}[t!]
\begin{center}
\begin{tabular}{l|ll|ll}
\hline \\
\parbox{1cm}{Model}&
\parbox{1.4cm}{\bf IDR (Instance)}&
\parbox{1.4cm}{\bf IFP (Instance)}&
\parbox{0.8cm}{\bf PDR (Pixel)}&
\parbox{0.8cm}{\bf PFP (Pixel)}\\
\\ \hline 
\\
Stripe Net@135 & 65.22 & 2.03 &74.65 & 1.98 \\
Context Net@135 & 55.89 & 0.49& 62.76 & 1.73\\ 
\bf MergeNet@135 & 73.42 &0.69& 85.00 & 2.01\\
Stripe Net@1036 & 77.87 & 2.48 & 86.71 & 4.16\\
Context Net@1036 & 65.00 & 1.40& 74.52 & 3.60\\ 
\bf MergeNet@1036 & 82.05 &0.76& 92.85 & 3.19\\
\hline 
\end{tabular}
\caption{MergeNet and its components' performance on the Lost and Found dataset}
\label{table:final_results}\vspace{-3.5em}
\end{center}
\end{table}
\subsection{Quantitative results}
Table \ref{table:final_results} presents the results of our model and all its components, on the full testing set of 1036 images and on a subset of it containing 135 images. The Stripe-net model performs better than the Context-net, reinforcing two of our basic assumptions. Firstly, that the road scenes afford an inherent vertical structure in the data, for both RGB and depth inputs, which can be exploited effectively by sharing parameters across the strips. Secondly, the local information learnt by the Stripe-net model is more capable of detecting the small obstacles than the standard architectures in semantic segmentation, which operate on full images. On the other hand, the Context-net is much more useful in maintaining the consistency of labels in a globally-aware manner, as evident from the qualitative results in the next section. Furthermore, the Context-net also plays the vital role of learning features which minimize the presence of false positives. This is also verified quantitatively by observing that the false positives are reduced from $2.03$ (Stripe-Net@135) to $0.69$ (Refiner-net@135) and from $2.48$ (Stripe-Net@1036) to $0.76$ (Refiner-net@1036)  when going from only Stripe-Net to the Refiner-Net (final model) which uses both Context and Stripe networks.

Table \ref{table:compare_results} shows the comparison of our model against the previous state-of-art methods. FPHT Stixels~\cite{Carsten-IROS16} uses the same 1200 image testing set from Lost and Found dataset that we do. They also have a similar detection metric for pixel-level and instance-level detection which makes a direct comparison feasible. Using only 135 images, we achieve an instance detection rate of $73.4\%$ which is a $19\%$ improvement with $10$ times less the data used for training and much fewer false positives. A similar improvement is also visible in pixel-level detection rates and false positives. Naturally, this performance boost is even higher (around $30\%$) when comparing our approach trained on the full training set with FPHT Stixels.

We also compare our model with a more recent work called UON-Stixels and its extensions~\cite{Carsten-NIPS17}  which are prefixed as \textit{FUSION} in table \ref{table:compare_results}. We perform comparable to their state-of-art detection result on instances with only one-sixth of the training data which again affirms the ability of MergeNet to perform better with lesser data (it is to be noted that we use a stricter metric of at least $50\%$ overlap with the ground truth annotation whereas their metric only evaluates on an upper bound). The other metrics like pixel-level detections and false positives can't be compared since they are either not evaluated in the paper or are differently defined. Another important point to be noted here is that UON Stixels and its extensions also learn a fully convolutional network trained for semantic segmentation for generating a semantic class map which is then used by other methods to get the $84.1\%$ detection rate. This means that it is theoretically possible to use our networks in conjunction with their post-processing techniques to further improve our performance.

  \begin{figure}[thpb]
     \centering
      \includegraphics[width=\linewidth]{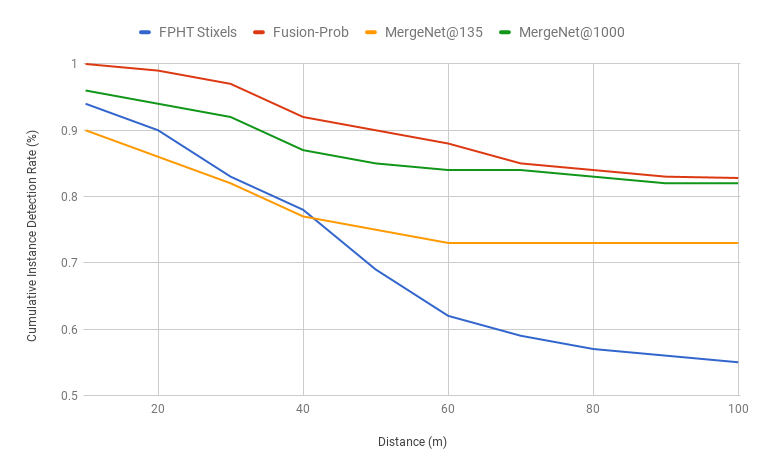}\vspace{-1em}
     \caption{Variation of the cumulative instance detection rate across obstacle distance}
     \label{fig:chart}
   \end{figure}

Another quantitative experiment which we conduct in Figure~\ref{fig:chart}, similar to that of previous work, is analyzing the variation of cumulative detection performance with respect to the depth of obstacle. The plot shows two of our models, along with the plots from the previous work. The graph presents the typical difficulty in detecting obstacles as the distance from the camera increases. 

Finally, we report the inference time of our model. MergeNet generates semantic maps at $5$ FPS on a Nvidia GeForce GTX 1080 Ti GPU which can be run smoothly on top of an autonomous vehicle in real time.

\begin{table}[t!]
\begin{center}
\begin{tabular}{l|ll|ll}
\hline \\
\parbox{1cm}{Model}&
\parbox{1.4cm}{\bf IDR  (Instance)}&
\parbox{1.4cm}{\bf IFP  (Instance)}&
\parbox{0.8cm}{\bf PDR (Pixel)}&
\parbox{0.8cm}{\bf PFP (Pixel)}\\
\\ \hline 
\\
FPHT Stixels@1000 & 55.00 &5.00& 68.00 & 2.00\\
UON-Stixels@6000 & 73.80 &0.103& NA & NA \\
FUSION-OR@6000 & 84.10 &0.669& NA & NA \\
FUSION-Prob@6000 & 82.80 &0.496& NA & NA \\
MergeNet@135 & 73.42 &0.69& 85.00 & 2.01\\
\bf MergeNet@1036 & 82.05 &0.76& 92.85 & 3.19\\
\hline 
\end{tabular}
\caption{Comparison of our method with ~\cite{Carsten-IROS16} and ~\cite{Carsten-NIPS17}}
\label{table:compare_results}\vspace{-3.5em}
\end{center}
\end{table}

\begin{figure*}[t!]
\centering
\begin{tabular}[b]{c c c}

\includegraphics[width=0.33\linewidth]{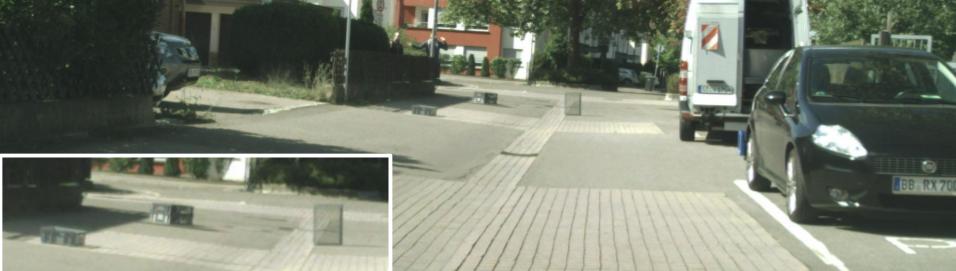}& \hspace{-1.2em}
\includegraphics[width=0.33\linewidth]{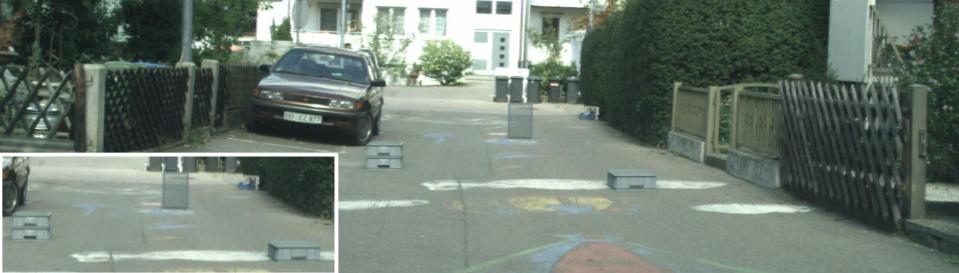}& \hspace{-1.2em}
\includegraphics[width=0.33\linewidth]{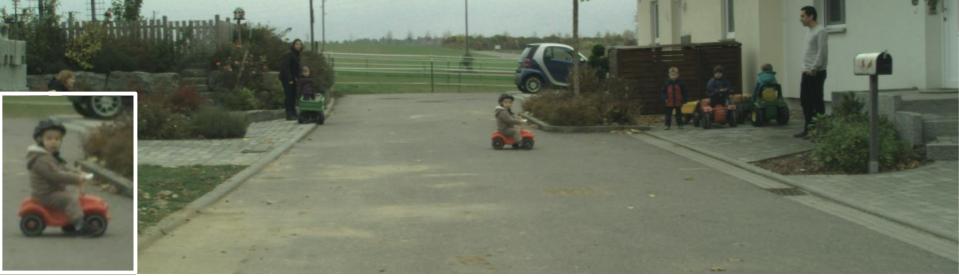} \vspace{0.1em}\\
{} & {Images from Lost \& Found dataset} & {} \vspace{0.7em}\\

\includegraphics[width=0.33\linewidth]{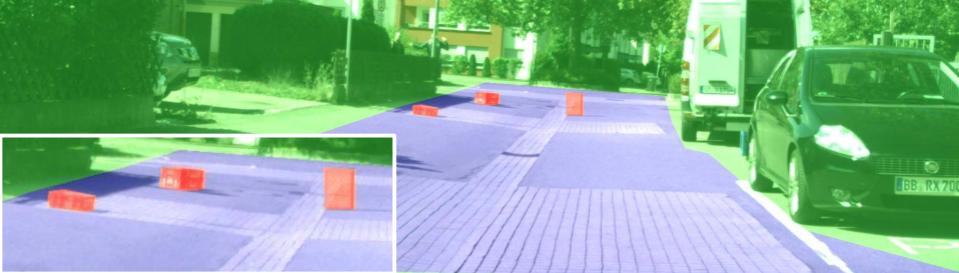}& \hspace{-1.2em}
\includegraphics[width=0.33\linewidth]{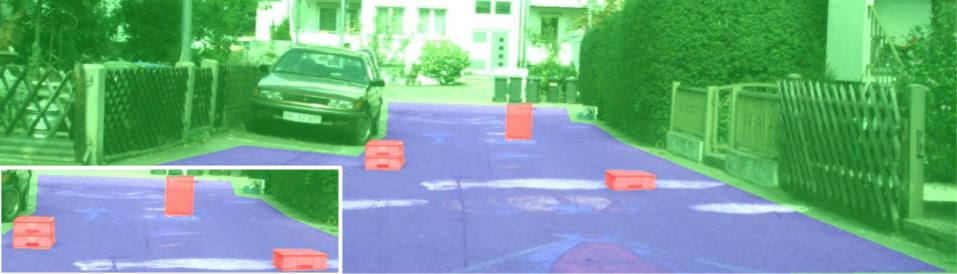}& \hspace{-1.2em}
\includegraphics[width=0.33\linewidth]{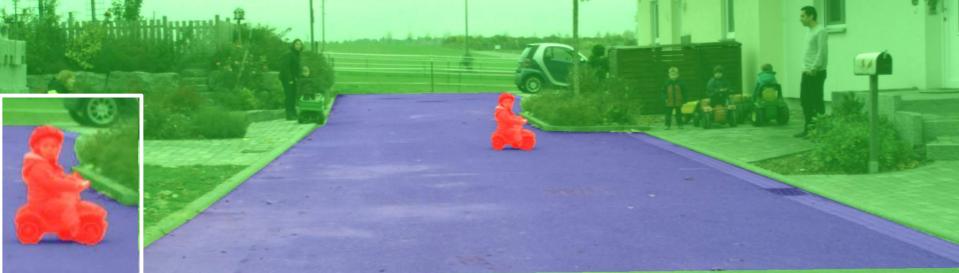} \vspace{0.1em}\\
{} & {Ground Truth} & {} \vspace{0.7em}\\

\includegraphics[width=0.33\linewidth]{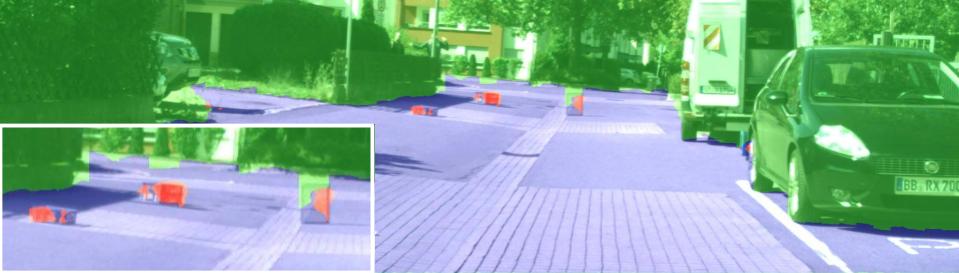}& \hspace{-1.2em}
\includegraphics[width=0.33\linewidth]{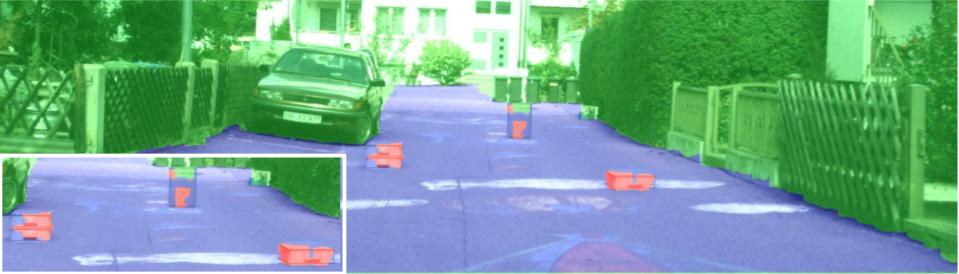}& \hspace{-1.2em}
\includegraphics[width=0.33\linewidth]{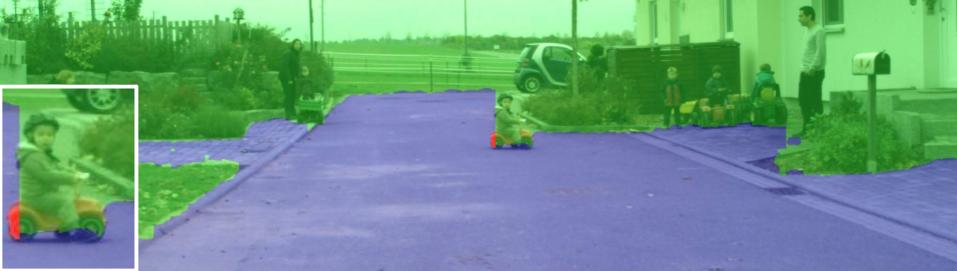} \vspace{0.1em}\\
{} & {Outputs of Stripe Net} & {} \vspace{0.7em}\\

\includegraphics[width=0.33\linewidth]{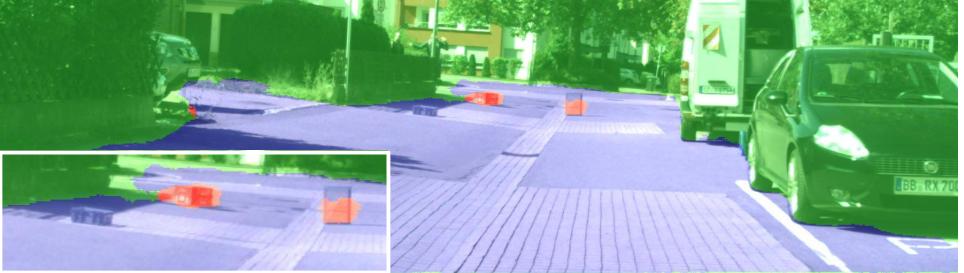}& \hspace{-1.2em}
\includegraphics[width=0.33\linewidth]{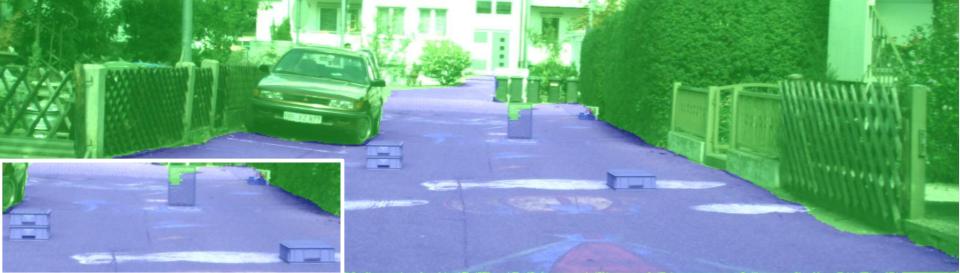}& \hspace{-1.2em}
\includegraphics[width=0.33\linewidth]{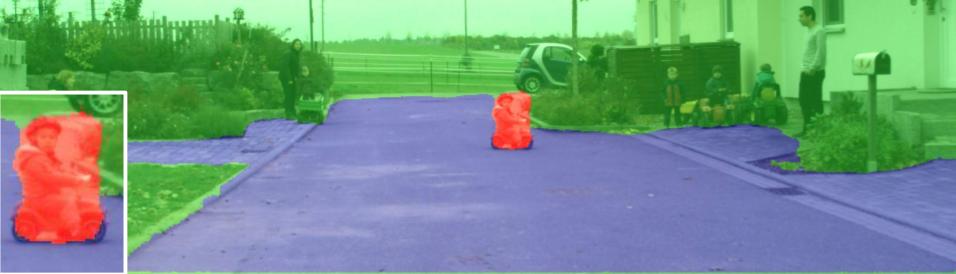} \vspace{0.1em}\\
{} & {Outputs of Context Net} & {} \vspace{0.7em}\\

\includegraphics[width=0.33\linewidth]{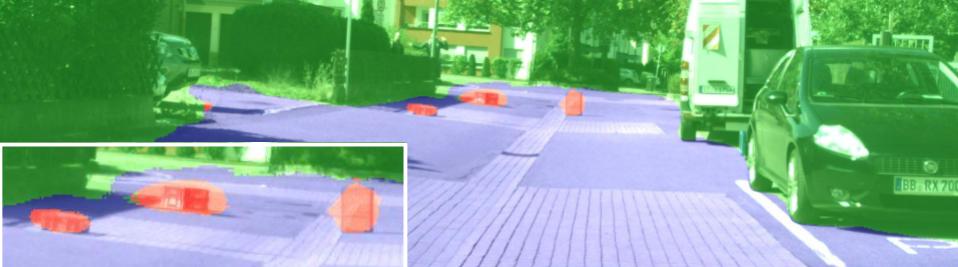}& \hspace{-1.2em}
\includegraphics[width=0.33\linewidth]{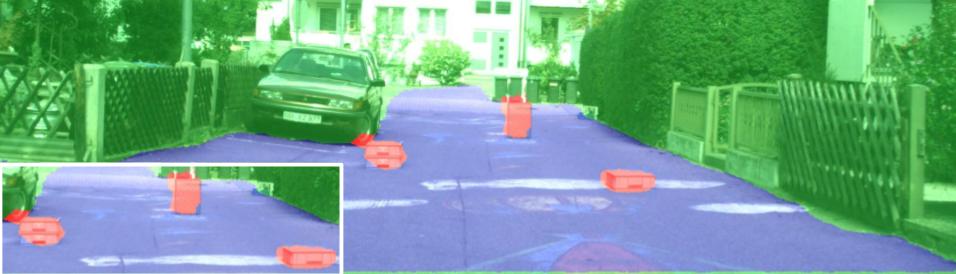}& \hspace{-1.2em}
\includegraphics[width=0.33\linewidth]{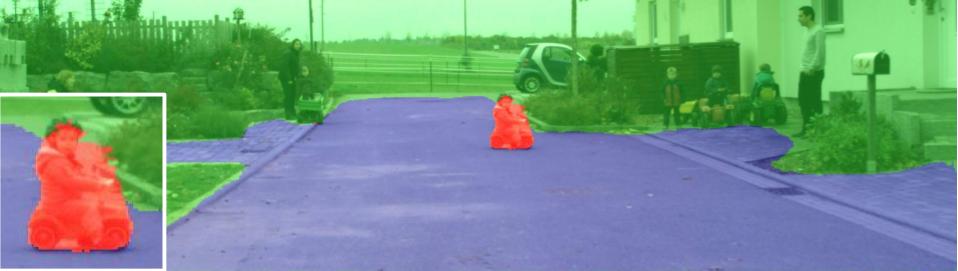} \vspace{0.1em}\\
{} & {Final output of MergeNet} & {}  \vspace{0.7em}\\

\end{tabular}\vspace{-1em}
\caption{Obstacle detection results on some of the selected examples using MergeNet@135. The obstacle category is marked in red, the off road category is marked in green and the road category is marked in blue. Please zoom in to clearly see the small obstacles (best viewed in color)}
\label{fig:result2}\vspace{-0.5em}
\end{figure*}

\subsection{Qualitative results}
Results of our method (MergeNet@135) on some of the images from the test set are shown in Figure~\ref{fig:result2}. Additional results can be found in our video. We select the particular road images to highlight the challenges arising due to varying appearance, size, distance, shape and clutter of the obstacles. The qualitative results display all the network outputs of 3 test images in each column. The first column shows three far off obstacles on the road, one of each going undetected by both stripe and context networks. But due to the complementary nature of features, and in turn, detections, the final refiner network output detects all three obstacles successfully. In the second column, three obstacles are present, each with different size, height, texture and illumination. This road also has chalk markings, something which the training set frames do not contain. Even though the context network misses out on all three obstacles, the stripe network detects all of them and this is also passed onto the refiner network, which segments all three objects with higher accuracy than both other networks. Finally, the third column shows a relatively larger obstacle, which is bigger than the strip-width of the stripe network. The stripe network merges it into the off road category, which is reasonable given it looks at each stripe individually. However, the prediction is corrected when merged with the context through the refiner network. 

As for the failure cases of our approach, we observe 3 main qualitative scenarios: a) Undetected obstacle - Occurs when the obstacle height is too low (e.g. a thin plank). b) Obstacle detected as off-road - Occurs when obstacle lies too close to the camera.
c) Off-road detected as obstacle - Occurs when irregular artifacts (e.g. patch of grass) are present at the boundary of the road. 


\section{Conclusion}
In this paper, we propose a novel deep network architecture for learning to detect small obstacles on the road scene. The model is composed of multiple stages, with each stage learning complementary features which are then fused to predict a segmentation map of the scene. We present thorough quantitative and qualitative experimentation and the results showcase high fidelity segmentation of obstacles on challenging public datasets. The current version of our algorithm runs at 5fps, which makes it suitable for on-road driving applications such as autonomous driving and driver assistant systems. In future work, we plan to investigate the proposed architecture for settings with more number of classes like pedestrian, pot holes, speed breakers and traffic signs.

\bibliographystyle{IEEEtran}
\bibliography{small_obstacle}

\end{document}